\documentclass{article}

\usepackage[eandd, final]{paper}

\makeatletter
\renewcommand{\@notice}{}
\makeatother

\usepackage[utf8]{inputenc}
\usepackage[T1]{fontenc}
\usepackage{hyperref}
\usepackage{url}
\usepackage{booktabs}
\usepackage{amsfonts}
\usepackage{amsmath}
\usepackage{nicefrac}
\usepackage{microtype}
\usepackage[table]{xcolor}
\usepackage{graphicx}
\usepackage{subcaption}
\usepackage{multirow}
\usepackage{arydshln}
\usepackage{wrapfig}

\graphicspath{{img/}}

\title{GeoR-Bench: Evaluating Geoscience\\ Visual Reasoning}

\author{
  \textbf{Yushuo Zheng\textsuperscript{1}},
  \textbf{Zicheng Zhang\textsuperscript{1,$\dagger$}},
  \textbf{Huiyu Duan\textsuperscript{1}},
  \textbf{Chunyi Li\textsuperscript{1}},
  \textbf{Zijian Chen\textsuperscript{1}},
  \textbf{Ziheng Jia\textsuperscript{1}},
\\
  \textbf{Yue Shi\textsuperscript{1}},
  \textbf{Ke Gu\textsuperscript{2}},
  \textbf{Xiongkuo Min\textsuperscript{1}},
  \textbf{Guangtao Zhai\textsuperscript{1}}
\\
\\
  \textsuperscript{1}Shanghai Jiao Tong University \quad \textsuperscript{2} Beijing University of Technology
}

\begin{document}
\maketitle

\renewcommand{\thefootnote}{$\dagger$}
\footnotetext{Project Lead.}
\renewcommand{\thefootnote}{\arabic{footnote}}
\setcounter{footnote}{0}

\begin{abstract}
Geoscience intelligence is expected to understand, reason about, and predict earth system changes to support human decision-making in critical domains such as disaster response, climate adaptation and environmental protection. Although current research has shown promising progress on specific geoscience tasks, such as remote sensing interpretation, geographic question-answering, existing benchmarks remain largely task-specific which failing to capture the open-ended real world geoscience problems. As a result, it remains unclear how far current AI systems are from achieving genuine geoscience intelligence.
To address this gap, we present \textbf{GeoR-Bench}, a \underline{Bench}mark for evaluating \underline{Geo}science visual \underline{R}easoning through reasoning informed visual editing tasks. GeoR-Bench contains 440 curated samples spanning 6 geoscience categories and 24 task types, covering earth observation imagery and structured scientific representations such as maps and diagrams.
We evaluate outputs along three dimensions, including reasoning, consistency, and quality. 
Benchmark results of 21 closed- and open-source multimodal models reveal that geoscience reasoning remains a critical bottleneck. The highest-performing model achieves 42.7\% overall strict accuracy, while the best open-source models only get 10.3\%. Notably, the visual consistency and image quality of the outputs frequently surpass their scientific accuracy. Ultimately, these findings indicate that current models generate superficially plausible results but fail to capture underlying earth science processes.
\end{abstract}

\section{Introduction}
\label{sec:introduction}

Geoscience intelligence is the capacity of an AI system to understand, reason about, and predict Earth-system changes in ways that can inform human decision-making in domains such as disaster response, climate adaptation, and environmental protection. Recent multimodal models have shown encouraging progress on narrow geoscience tasks, including remote sensing interpretation and geographic question answering. Yet a central question remains open, \textit{i.e.}, how do we \emph{evaluate} geoscience intelligence in a way that reflects the open-ended, visually grounded nature of real-world Earth-system problems? Existing benchmarks remain largely task-specific and measure isolated recognition or captioning abilities, leaving it unclear how far current models are from genuine geoscience intelligence. In practice, geoscience reasoning must operate over intermediate visual artifacts, including satellite images, maps, cross-sections, and scientific diagrams, and produce outputs that are not merely plausible but scientifically valid. A principled evaluation of geoscience intelligence must therefore probe whether a model can convert Earth-system understanding into a correct, visually grounded generation, rather than only verifying surface-level visual realism or textual recall.

To make geoscience intelligence measurable, we propose to evaluate it through reasoning-informed visual editing: given an input image and a task instruction, a model must infer the relevant Earth-system state and produce an output image that reflects that inference. This setting is challenging because correctness depends on reasoning about process, geometry, scale, topology, and domain conventions at the same time; predicting delta growth requires understanding sediment transport and channel branching, modifying a cyclone image requires respecting circulation structure, and completing a tectonic diagram requires correct labels and an interpretable layout. Pure recognition benchmarks do not force a model to externalize such understanding in a testable visual artifact, while generic generation benchmarks reward plausibility or instruction following without verifying scientific validity. Reasoning-informed visual editing closes this gap: the edited image becomes direct evidence of whether the model can turn geoscience reasoning into a grounded visual generation, making editing the measurement interface through which geoscience intelligence becomes observable.

\begin{figure*}[t]
    \centering
    \includegraphics[width=\textwidth]{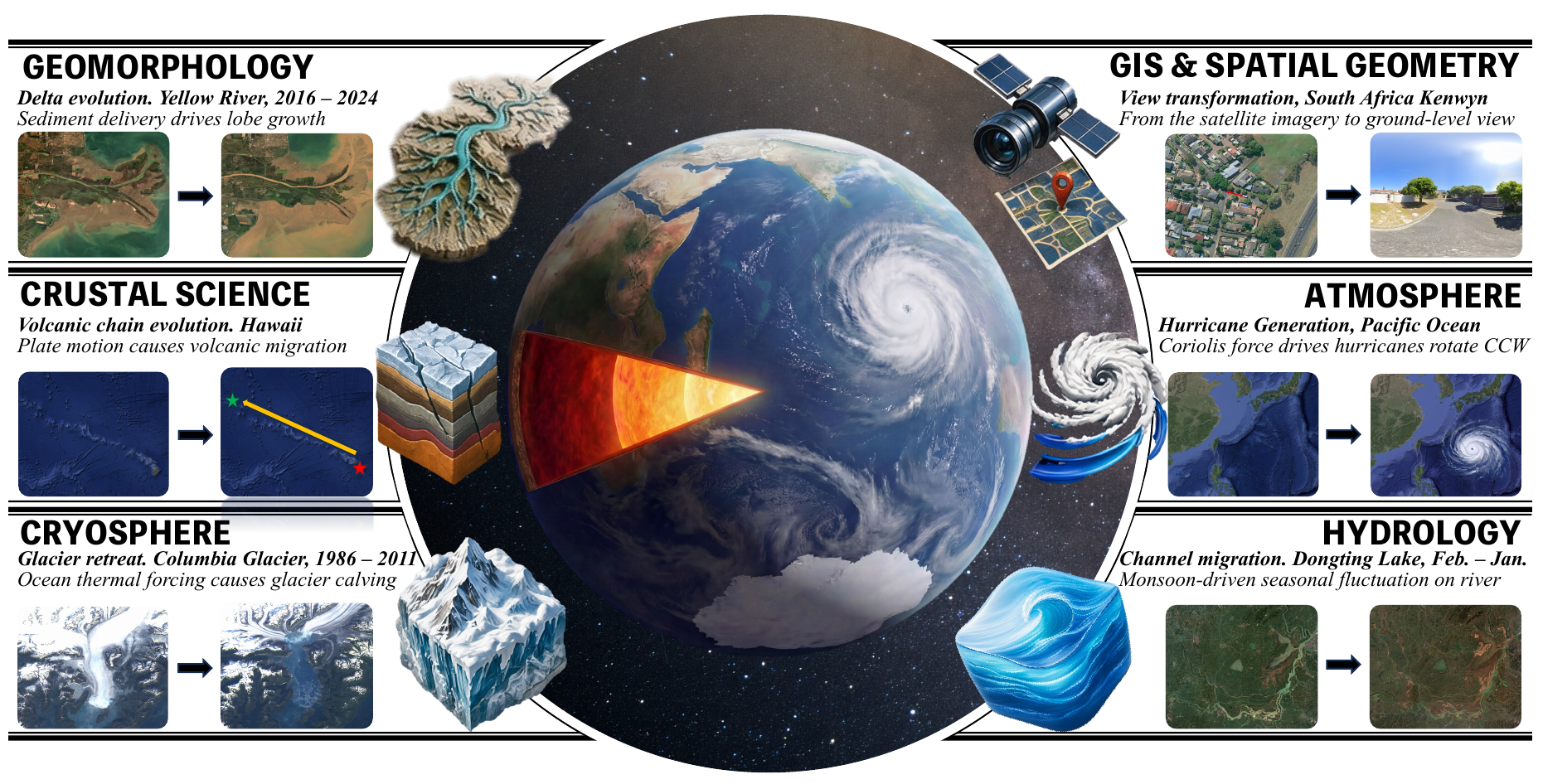}
    \caption{\textbf{GeoR-Bench} evaluates geoscience intelligence via reasoning-informed visual editing across six categories: geomorphology, GIS \& spatial geometry, crustal science, atmosphere, cryosphere, and hydrology. Each task requires inferring the underlying Earth-system process from an input image and producing a scientifically valid output.}
    \label{fig:teaser}
\end{figure*}

However, existing benchmarks do not directly evaluate this capability. General image-editing benchmarks emphasize natural-image manipulation, instruction following, and broad visual reasoning \cite{yang2025imgedit}. More reasoning-aware suites such as KRIS-Bench and RISEBench move beyond low-level editing quality, but they still focus primarily on natural scenes and generic structural reasoning rather than scientific correctness \citep{wu2025krisbench,xia2025risebench}. Scientific and multimodal knowledge benchmarks such as MMMU evaluate expert-level understanding \citep{yue2024mmmu}, while scientific illustration benchmarks test technical drawing ability \citep{chang2025srdbench}. These works are important, but they rarely require a model to express domain understanding through a visually grounded geoscience generation. Therefore, in this work, we address this gap by evaluating whether multimodal models can convert geoscience understanding into a correct edited or generated image.

GeoR-Bench is built around this principle, which contains 440 samples spanning 6 categories and 24 task types, with 4 task types per category. The benchmark covers geomorphology, hydrology, atmosphere and ocean dynamics, cryosphere science, GIS and spatial geometry, and crustal science. Its tasks span both real-world Earth observation imagery and formal scientific representations, including river delta evolution, wetland fluctuation, cyclone generation under Coriolis constraints, snowline prediction, cross-view geospatial reconstruction, earthquake epicenter triangulation, and geological labeling. This breadth is deliberate: geoscience visual reasoning must operate across natural imagery, spatial products, and structured scientific diagrams rather than within a single visual modality. Figure~\ref{fig:teaser} illustrates the benchmark scope.

To evaluate model outputs, we use a three part protocol to judge the generated images: \textbf{reasoning}, \textbf{consistency}, and \textbf{quality}. Reasoning evaluates whether the generated output reflects the intended physical, spatial, or process-based generation by jointly comparing the input image, ground-truth output, model prediction, and the task-specific rubric. Consistency evaluates whether the generated output preserves the required visual relationship between the input image and the target output. Quality evaluates whether the generated image is visually usable as an artifact. We also report \emph{accuracy}, which counts a sample as correct only when all three dimensions are satisfied simultaneously. These metrics support claims about benchmark-defined geoscience visusal resosing ability.

Our contributions are as follows:
\begin{itemize}
\item We articulate a comprehensive evaluation principle for geoscience intelligence: it should be tested through reasoning-informed visual editing tasks that require models to convert Earth-system understanding into scientifically correct visual outputs.

\item We introduce GeoR-Bench, a benchmark of 440 curated samples across 6 categories and 24 task types. Each sample follows a standardized format, each task has a dedicated judging rubric, and all evaluative claims are clearly documented.

\item We provide experiments and analysis showing that current 21 multi-modal models remain substantially weaker at geoscience reasoning than at preserving visual appearance or producing readable outputs, yielding concrete evidence about the present limits of multimodal geoscience visual reasoning. 
\end{itemize}

\section{Related Work}
\label{sec:related_work}

\subsection{Large Language Models and Multimodal Benchmarks}
Recent progress in multimodal large models has produced a new generation of evaluation benchmarks for image editing, instruction-following generation, and visually grounded reasoning. General-purpose editing benchmarks such as ImgEdit assess instruction alignment, local content preservation, and visual plausibility across diverse natural scenes \citep{yang2025imgedit}, but strong performance there does not imply fidelity to deeper causal or domain constraints. Reasoning-oriented benchmarks including KRIS-Bench and RISEBench show that visually attractive edits often violate temporal, spatial, causal, or logical requirements \citep{wu2025krisbench,xia2025risebench}, while WorldGenBench tests whether text-to-image systems convert implicit world knowledge into faithful outputs \citep{cao2025worldgenbench}, and broader knowledge benchmarks such as MMMU and SrdBench highlight persistent difficulty with expert knowledge and technical visual conventions \citep{yue2024mmmu,chang2025srdbench}. Yet these benchmarks center on natural scenes or generic knowledge, leaving untested the geoscience-specific combination of Earth-system process, spatial geometry, scale consistency, and disciplinary convention, an output may look plausible while misrepresenting channel branching, cyclone structure, or crustal labels. GeoR-Bench fills this gap by evaluating whether large models can convert domain understanding into scientifically correct visual outputs.

\subsection{Geoscience Benchmarks and Evaluation}
Within remote sensing and Earth science, existing benchmarks have expanded multimodal evaluation but emphasize different capabilities. One line of work targets visual understanding of Earth observation imagery: VRSBench benchmarks remote-sensing vision-language models on captioning, grounding, and question answering \citep{li2024vrsbench}, while multimodal datasets such as MDAS combine optical, hyperspectral, SAR, DSM, and GIS sources to support richer perception tasks \citep{hu2023mdas}. A second line moves toward generation and scientific reasoning: MMM-RS extends remote-sensing benchmarking to text-to-image generation across scenes, modalities, weather, and resolutions \citep{luo2024mmmrs}, and MSEarth and GRADE curates scientific figures and enriched reasoning content from the literature for multimodal scientific understanding \citep{zhao2025msearth, liu2026gradebenchmarkingdisciplineinformedreasoning}. However, none of these directly tests whether a model can start from an \emph{input} geoscience image, infer the required Earth-system change, and produce an \emph{output} image that remains scientifically correct under task-specific constraints. GeoR-Bench is therefore neither a pure remote-sensing understanding benchmark nor a pure text-to-image generation benchmark. It covers both real-world Earth observation imagery and formal representations such as maps, cross-sections, and diagrams, it complements prior work while targeting a capability they do not measure.

\section{The GeoR-Bench Benchmark}
\label{sec:benchmark}
\begin{figure*}[t]
    \centering
    \includegraphics[width=\textwidth]{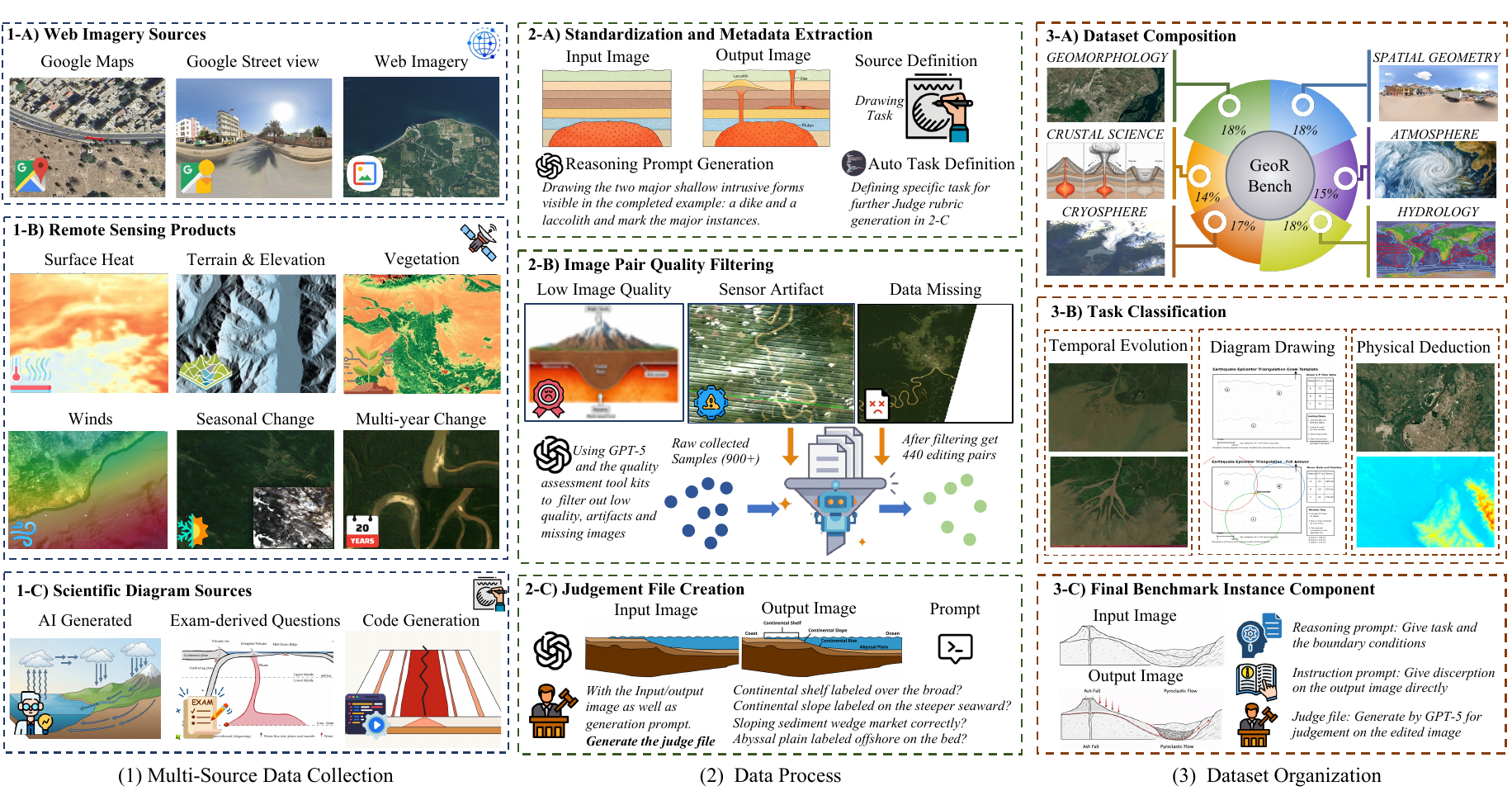}
    \caption{Overview of the GeoR-Bench. The GeoR-Bench is composed by multi-scource geoscience images including the Web Imagery, Remote Sensing, Scientific Diagrams, and Map Products. The raw data has been filtered and reorganized into a standardized format for model evaluation. The dataset is balanced across six geoscience categories,with unique dedicated judging rubric.}
    \label{fig:overview}
\end{figure*}

\subsection{Principle \& Motivation}
The central principle of GeoR-Bench is that geoscience visual reasoning should be evaluated through scientific tasks that require both understanding and visual generation. Image editing is not the target capability by itself; it is the observable interface through which a model must reveal whether it understands Earth-system structure, process, geometry, and scientific convention. A model succeeds only when it can interpret a geoscience image, infer the required generation, and generate an output image that is scientifically correct, turning image reasoning-and-generation into a benchmark of geoscience reasoning rather than a test of aesthetic plausibility alone.

Geoscience is a particularly demanding target for this framing because many tasks are process-based rather than appearance-based. The model must infer why a delta grows, why a storm rotates, where a snowline should sit, or how a tectonic cross-section should be labeled. All of those task are multi-scale, requiring local geometric edits, global spatial consistency, and physically meaningful background preservation within a single example. The domain also spans heterogeneous visual forms, from natural-looking satellite imagery to scientific cross-sections, topographic renderings, and cartographic products, so a benchmark that claims to measure geoscience competence must cover both real-world imagery and formal scientific representations. Within this scope, 

This framing motivates GeoR-Bench as a purpose-built probe for geoscience visual reasoning. Existing image-editing benchmarks largely reward photorealism or local fidelity, which leaves a blind spot precisely where Earth-system competence lives: whether a model can read a geoscience scene, reason about the underlying process, and render an output that is scientifically defensible under domain-specific criteria. Our goal is to turn image generation into a measurable act of geoscience reasoning, so that progress on the benchmark translates into progress on the scientific understanding.

\begin{figure*}[t]
    \centering
    \includegraphics[width=\textwidth]{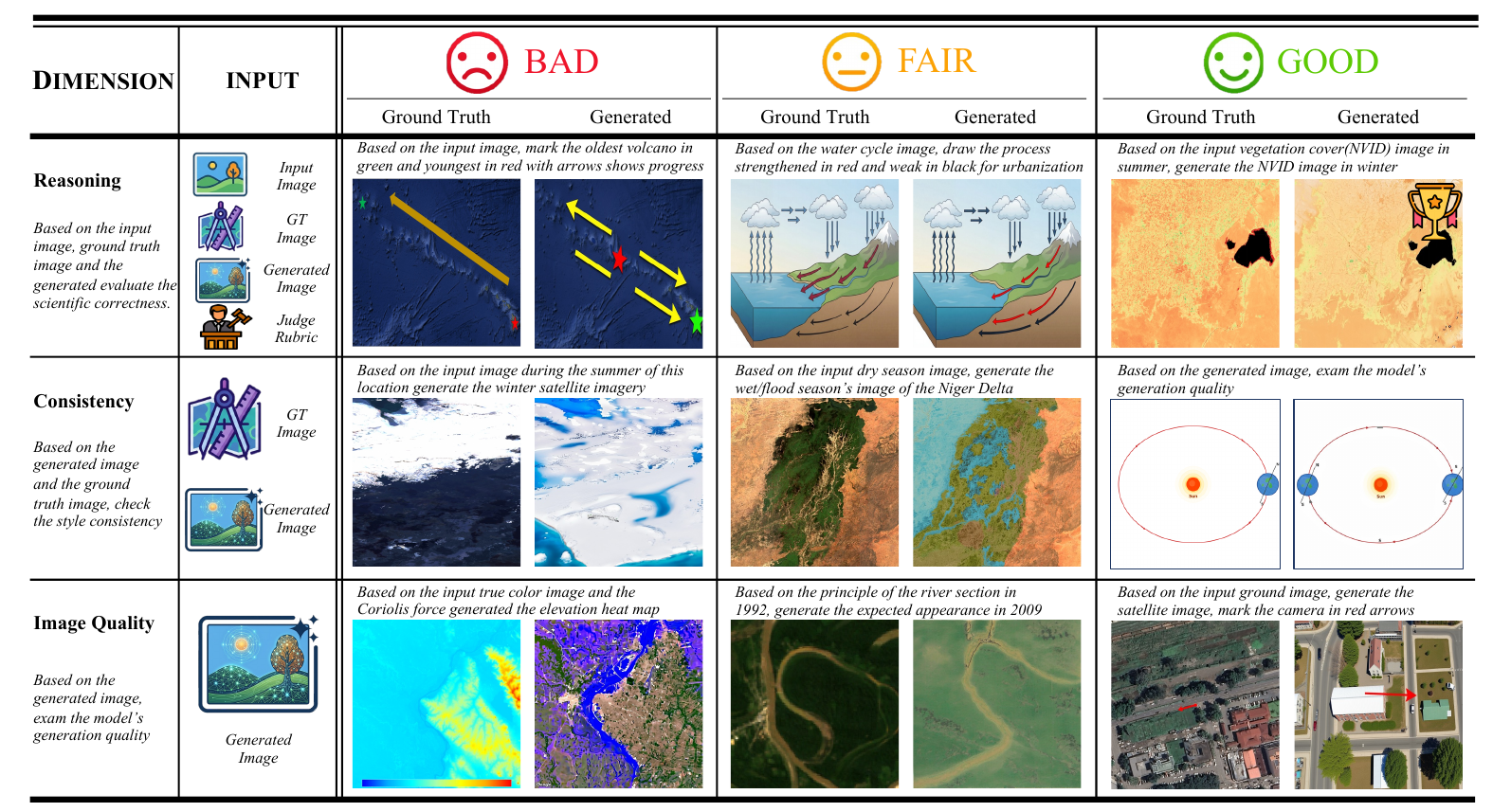}
    \caption{GeoR-Bench evaluation pipeline jointly evaluates reasoning consistency and quality. Reasoning uses the input ground-truth model prediction and the judge rubric to evaluate correctness. Consistency assesses visual agreement while quality measures the final generated output.
}
    \label{fig:pipeline}
\end{figure*}

\subsection{Dataset Construction \& Composition}
GeoR-Bench is built around a unified sample format while preserving raw sources. The benchmark is primarily composed of domain curated geoscience images selected for clear visual structure and task suitability, as shown in Figure~\ref{fig:overview}. In practice, these include satellite observations, map products, DEM renderings, scientific cross-sections, and structured geoscience diagrams. The release contains 440 standardized samples spanning 6 categories and 24 task types, organized so that each category contains 4 task types. The release spans the following categories: geomorphology, hydrology, atmosphere and ocean dynamics, cryosphere science, GIS and spatial geometry, and crustal science. This balanced category structure is intentional: the benchmark is designed to cover the breadth of geoscience visual reasoning rather than over-concentrating on any single modality or subfield.

Within this structure, the benchmark includes both real-world Earth observation tasks and structured scientific-figure tasks. Representative task types include river delta evolution, oxbow lake formation, monsoon lake wet to dry transition, cyclone generation and transposition, glacier retreat, snowline altitude prediction, crater generation, street-to-satellite and satellite-to-panorama reconstruction, earthquake epicenter triangulation, seismogram interpretation, volcanic plumbing diagrams, and marine geology labeling. This mixture is deliberate. Geoscience competence is not only about satellite forecasting, and it is not only about textbook diagrams. The geoscience intelligence should performance well for both visual regimes to produce reliable results.

To evaluate models uniformly across these heterogeneous sources, each sample is reorganized into a common form: an input image, a ground-truth output image, a natural-language prompt, a task-specific judging rubric, and, when available, auxiliary context images and structured metadata. A single benchmark index enumerates all samples, and a unified dataloader exposes each one as a consistent evaluation unit regardless of its raw origin. This standardization makes it possible to mix raw source patterns, such as temporal satellite series, cross-view geospatial data, metadata-driven tasks, and structured figure-editing tasks, within one evaluation interface.

\subsection{Evaluation Protocol}

Our evaluation pipeline is designed specifically for geoscience image editing and generation, as shown in Figure~\ref{fig:pipeline}. To ensure generated images are practically useful for Earth system analysis, our framework assesses outputs through a progression of four interrelated dimensions. First, we assess Reasoning to determine whether the underlying scientific reasoning-informed visual editing is physically and domain accurately executed. We then measure Consistency to verify that the required spatial, structural, and stylistic relationships with the source input are preserved. Next, we evaluate the foundational Quality of the artifact to ensure it is free of basic visual degradations. Finally, the strict accuracy metric unifies these independent axes and requires simultaneous success across all dimensions to demonstrate genuine geoscience intelligence.

\paragraph{Reasoning.}
Reasoning evaluates whether the model performs the intended geoscience generation correctly. For each sample, the evaluator takes the input image, the ground-truth output image, the model prediction, and the task-specific \texttt{judge.md} rubric. The rubric defines scoring points aligned with the scientific objective of the task. For example, a delta-evolution task may score whether the delta front progrades seaward and whether distributary channels bifurcate correctly, while an epicenter task may score whether the circles are drawn at the right radii and intersect at the correct point. The weighted sum yields a reasoning score in $[0,1]$, then been scaled in $[0,100]$.

\paragraph{Consistency.}
Consistency evaluates the required visual relationship between the generated output, the input image, and the ground-truth target. For each sample, the evaluator takes the input image, the ground-truth output image, and the model prediction, and measures whether the prediction preserves the visual relationship implied by the task. In style-sensitive tasks, this means preserving the sensor or rendering style of the source while matching the target visual appearance. In localized tasks, consistency emphasizes preserving content outside the instructed edit. In independent-output tasks, consistency verifies that the generated output remains visually compatible with the expected target representation rather than identical in style to the input.

\paragraph{Quality.}
Quality evaluates the generated output using an image-quality assessment method applied to the model prediction itself. This component measures whether the generated image is visually well formed as an output artifact, independent of whether the underlying geoscience reasoning is correct. It captures degradations such as blur, distortion, artifacts, and other failures that reduce the usability of the generated result.

\paragraph{Strict accuracy.}
We report a strict accuracy metric that counts a sample as correct only when Reasoning, Consistency, and Quality are simultaneously satisfied. This metric is intentionally demanding: in geoscience practice, an image that is visually clean but scientifically wrong, or scientifically plausible but inconsistent with the required representation, is still a failed output.

\section{Experiments}
\label{sec:experiments}

\begin{table*}[t]\small
\caption{Reasoning and Consistency result across 21 models, with the best score in each column highlighted in \textbf{bold} and the second- and third-best scores \underline{underlined}.} 
\centering
\renewcommand\arraystretch{1.02}
\setlength{\tabcolsep}{6pt}
\resizebox{\linewidth}{!}{%
\begin{tabular}{l|cccccccccccc|cc}
\hline\hline
\multirow{2}{*}{\textbf{Model}} & \multicolumn{14}{c}{\textbf{Evaluation Dimensions}} \\
\cline{2-15}
& \multicolumn{2}{c}{\textit{GEOM}} & \multicolumn{2}{c}{\textit{HYDR}} & \multicolumn{2}{c}{\textit{ATMO}} & \multicolumn{2}{c}{\textit{CRYO}} & \multicolumn{2}{c}{\textit{SPATIAL}} & \multicolumn{2}{c|}{\textit{CRUSTAL}} & \multicolumn{2}{c}{\textit{OVERALL}} \\
\cdashline{2-3}\cdashline{4-5}\cdashline{6-7}\cdashline{8-9}\cdashline{10-11}\cdashline{12-13}\cdashline{14-15}
& \textit{R} & \textit{C} & \textit{R} & \textit{C} & \textit{R} & \textit{C} & \textit{R} & \textit{C} & \textit{R} & \textit{C} & \textit{R} & \textit{C} & \textit{R} & \textit{C} \\
\hline
\multicolumn{15}{l}{\textit{Closed Source Models}} \\
\cdashline{1-15}
FLUX.2 Max~\citep{flux-2-2025} & 37.2 & 65.0 & 57.8 & 68.8 & 49.4 & 82.4 & 45.3 & 72.9 & 47.1 & 81.0 & 53.9 & 59.7 & 48.5 & 71.6 \\
\rowcolor{gray!10} FLUX.2 Pro~\citep{flux-2-2025} & 38.9 & 56.2 & 46.4 & 67.5 & 42.5 & 72.1 & 35.1 & 51.4 & 36.1 & 59.5 & 56.5 & 61.3 & 42.6 & 61.3 \\
GPT-Image-1.5~\citep{GPT-Image-1.5} & 56.9 & 88.8 & 74.5 & 93.8 & \underline{66.9} & \underline{97.1} & \underline{71.3} & \textbf{97.1} & 61.0 & \underline{92.9} & 78.2 & \underline{96.8} & 68.1 & 94.4 \\
\rowcolor{gray!10} GPT-Image-2~\citep{GPT-Image-2.0} & \textbf{71.7} & \textbf{95.0} & \underline{88.9} & \underline{96.2} & \textbf{72.4} & 92.6 & \textbf{82.4} & \underline{95.7} & \textbf{69.2} & \underline{94.0} & \textbf{90.0} & \underline{96.8} & \textbf{79.1} & \underline{95.1} \\
Nano Banana~\citep{Nano-Banana} & 49.2 & \underline{90.0} & 59.9 & 92.5 & 42.8 & 88.2 & 58.0 & \underline{94.3} & 51.1 & 76.2 & 63.1 & 87.1 & 54.0 & 88.1 \\
\rowcolor{gray!10} Nano Banana 2~\citep{Nano-Banana-2} & \underline{71.6} & 85.0 & \underline{86.9} & \textbf{100} & \underline{69.3} & \underline{94.1} & \underline{75.7} & \underline{94.3} & \underline{68.7} & \underline{94.0} & \underline{85.0} & \textbf{100} & \underline{76.2} & \underline{94.6} \\
Nano Banana Pro~\citep{Nano-Banana-Pro} & \underline{58.6} & \underline{91.2} & \textbf{90.6} & \underline{98.8} & 62.2 & \underline{97.1} & 66.7 & \underline{95.7} & \underline{65.5} & \textbf{96.4} & \underline{84.8} & \underline{98.4} & \underline{71.4} & \textbf{96.3} \\
\rowcolor{gray!10} Seedream 4.0~\citep{chen2025seedream4} & 45.0 & 52.5 & 36.0 & 77.5 & 48.4 & 80.9 & 43.3 & 55.7 & 22.4 & 47.6 & 37.4 & 35.5 & 38.7 & 58.3 \\
Seedream 4.5~\citep{Seedream4-5} & 50.0 & 68.8 & 55.1 & 72.5 & 45.1 & 82.4 & 51.1 & 81.4 & 29.0 & 63.1 & 48.7 & 59.7 & 46.5 & 71.3 \\
\rowcolor{gray!10} Seedream 5.0~\citep{Seedream5-0} & 50.2 & 81.2 & 76.0 & 88.8 & 56.5 & \textbf{98.5} & 58.3 & \underline{94.3} & 43.1 & 90.5 & 71.0 & 88.7 & 59.2 & 90.3 \\
\hline
\multicolumn{15}{l}{\textit{Open Source Models}} \\
\cdashline{1-15}
Bagel~\citep{deng2025bagel} & 10.1 & 22.5 & 16.5 & 28.8 & 20.7 & 20.6 & 12.3 & 18.6 & 10.6 & 25.0 & 5.81 & 21.0 & 12.7 & 22.7 \\
\rowcolor{gray!10} Bagel (think)~\citep{deng2025bagel} & 19.8 & 53.8 & 21.4 & 51.2 & 24.9 & 45.6 & 18.1 & 38.6 & 17.6 & 40.5 & 10.6 & 37.1 & 18.7 & 44.5 \\
DreamOmni~\citep{xia2025dreamomni} & 26.1 & 77.5 & 31.8 & 65.0 & 22.5 & 32.4 & 28.9 & 75.7 & 18.2 & 54.8 & 12.3 & 40.3 & 23.3 & 57.6 \\
\rowcolor{gray!10} FLUX.2 dev~\citep{flux-2-2025} & 33.0 & 75.0 & 34.1 & 80.0 & 37.4 & 77.9 & 41.4 & 52.9 & 21.1 & 44.0 & 34.4 & 43.5 & 33.6 & 62.2 \\
ICEdit~\citep{zhang2025icedit} & 18.6 & 42.5 & 12.1 & 33.8 & 21.8 & 41.2 & 11.1 & 37.1 & 13.3 & 28.6 & 2.42 & 3.23 & 13.2 & 31.1 \\
\rowcolor{gray!10} ICEdit (MoE)~\citep{zhang2025icedit} & 14.5 & 33.8 & 5.12 & 5.00 & 4.41 & 13.2 & 5.00 & 25.7 & 9.40 & 29.8 & 3.87 & 1.61 & 7.05 & 18.2 \\
OmniGen~\citep{xiao2025omnigen} & 1.75 & 6.25 & 5.00 & 3.75 & 16.6 & 8.82 & 4.14 & 10.0 & 1.19 & 17.9 & 0.323 & 6.45 & 4.84 & 8.86 \\
\rowcolor{gray!10} Qwen-Edit-2511~\citep{wu2025qwenimage} & 44.0 & 60.0 & 57.9 & 71.2 & 44.1 & 67.6 & 51.1 & 68.6 & 31.2 & 32.1 & 65.2 & 75.8 & 48.9 & 62.6 \\
Step-1x~\citep{han2025stepx} & 31.1 & 73.8 & 23.0 & 60.0 & 25.3 & 44.1 & 31.4 & 55.7 & 28.7 & 41.7 & 8.55 & 27.4 & 24.7 & 50.4 \\
\rowcolor{gray!10} Step-1x (think)~\citep{han2025stepx} & 22.6 & 66.2 & 24.9 & 55.0 & 27.1 & 42.6 & 30.9 & 54.3 & 23.5 & 22.6 & 9.19 & 35.5 & 23.0 & 46.0 \\
Step-1x (think+reflect)~\citep{han2025stepx} & 23.0 & 56.2 & 20.6 & 57.5 & 29.0 & 35.3 & 24.6 & 45.7 & 15.4 & 27.4 & 15.5 & 32.3 & 21.3 & 42.4 \\
\hline
\end{tabular}%
}
\label{tab:benchmark_rc_overall}
\end{table*}

\subsection{Experimental Setup}

We evaluate 21 image editing and image generation systems spanning the current closed and open source landscape. Those models are categorized into two groupds: (1)closed source model, including Nano Banana Pro~\citep{Nano-Banana-Pro}, Nano Banana 2~\citep{Nano-Banana-2}, Seedream 5.0~\citep{Seedream5-0}, GPT-Image-1.5~\citep{GPT-Image-1.5}, FLUX.2 Max~\citep{flux-2-2025}, Nano Banana~\citep{Nano-Banana}, Seedream 4.5~\citep{Seedream4-5}, GPT-Image-2~\citep{GPT-Image-2.0}, FLUX.2 Pro~\citep{flux-2-2025},Seedream 4.0~\citep{chen2025seedream4}
(2)open source model including Qwen-Edit-2511~\citep{wu2025qwenimage}, Step-1x~\citep{han2025stepx}, DreamOmni~\citep{xia2025dreamomni}, Bagel~\citep{deng2025bagel},ICEdit~\citep{zhang2025icedit}, FLUX.2 dev~\citep{flux-2-2025}, OmniGen~\citep{xiao2025omnigen}

Closed-source models are accessed directly through official APIs, while all open source models are deployed locally on NVIDIA H200 GPUs using their default inference configurations. Following the GeoR-Bench protocol, each model receives the standardized input image, prompt, and any task defined context images to generate a single predicted output. These generated outputs are then evaluated using Gemini 3 Flash\citep{Gemini3-Flash} and scored on reasoning, consistency using a normalized 0 to 100 scale.While using the Q-Align \citep{wu2023qalign} to evaluate the image quality. The main evaluation metric is strict accuracy, which requires a generated sample to simultaneously pass the reasoning, consistency, and quality dimensions.

\subsection{Main Results}
\begin{table*}[t]\small
\caption{Quality and Accuracy result across 21 models, with the best score in each column highlighted in \textbf{bold} and the second- and third-best scores \underline{underlined}.}
\centering
\renewcommand\arraystretch{1.02}
\setlength{\tabcolsep}{6pt}
\resizebox{\linewidth}{!}{%
\begin{tabular}{l|cccccccccccc|cc}
\hline\hline
\multirow{2}{*}{\textbf{Model}} & \multicolumn{14}{c}{\textbf{Evaluation Dimensions}} \\
\cline{2-15}
& \multicolumn{2}{c}{\textit{GEOM}} & \multicolumn{2}{c}{\textit{HYDR}} & \multicolumn{2}{c}{\textit{ATMO}} & \multicolumn{2}{c}{\textit{CRYO}} & \multicolumn{2}{c}{\textit{SPATIAL}} & \multicolumn{2}{c|}{\textit{CRUSTAL}} & \multicolumn{2}{c}{\textit{OVERALL}} \\
\cdashline{2-3}\cdashline{4-5}\cdashline{6-7}\cdashline{8-9}\cdashline{10-11}\cdashline{12-13}\cdashline{14-15}
& \textit{Q} & \textit{A} & \textit{Q} & \textit{A} & \textit{Q} & \textit{A} & \textit{Q} & \textit{A} & \textit{Q} & \textit{A} & \textit{Q} & \textit{A} & \textit{Q} & \textit{A} \\
\hline
\multicolumn{15}{l}{\textit{Closed Source Models}} \\
\cdashline{1-15}
FLUX.2 Max~\citep{flux-2-2025} & 69.7 & 2.50 & 77.3 & 10.0 & 79.3 & 14.7 & 74.0 & 5.71 & 73.0 & 19.0 & \textbf{81.6} & 9.68 & 75.8 & 10.3 \\
\rowcolor{gray!10} FLUX.2 Pro~\citep{flux-2-2025} & \underline{77.3} & 0.00 & \textbf{79.7} & 2.50 & \underline{80.9} & 5.88 & \textbf{77.4} & 8.57 & \textbf{79.4} & 7.14 & 79.8 & 9.68 & \textbf{79.1} & 5.63 \\
GPT-Image-1.5~\citep{GPT-Image-1.5} & 61.1 & 7.50 & 71.6 & 22.5 & 75.0 & \underline{23.5} & 71.7 & \underline{34.3} & 71.2 & 33.3 & 80.5 & \underline{35.5} & 71.9 & 26.1 \\
\rowcolor{gray!10} GPT-Image-2~\citep{GPT-Image-2.0} & 62.3 & \underline{20.0} & 72.7 & \underline{52.5} & 78.0 & \textbf{35.3} & 70.1 & \textbf{54.3} & 68.8 & \textbf{52.4} & 79.1 & \underline{41.9} & 71.8 & \textbf{42.7} \\
Nano Banana~\citep{Nano-Banana} & 61.4 & 5.00 & 72.8 & 22.5 & 76.2 & 2.94 & 69.1 & 25.7 & 71.4 & 11.9 & 79.0 & 19.4 & 71.7 & 14.6 \\
\rowcolor{gray!10} Nano Banana 2~\citep{Nano-Banana-2} & 69.6 & \textbf{25.0} & 77.3 & \underline{50.0} & 78.0 & \underline{26.5} & 75.5 & \underline{40.0} & 71.6 & \underline{45.2} & 80.5 & \textbf{48.4} & 75.4 & \underline{39.2} \\
Nano Banana Pro~\citep{Nano-Banana-Pro} & 56.9 & \underline{20.0} & 75.8 & \textbf{57.5} & 76.0 & 20.6 & 71.8 & \underline{34.3} & 71.8 & \underline{35.7} & 79.7 & \underline{41.9} & 72.0 & \underline{35.0} \\
\rowcolor{gray!10} Seedream 4.0~\citep{chen2025seedream4} & 75.8 & 5.00 & 75.5 & 2.50 & \underline{80.5} & 5.88 & 74.2 & 8.57 & 74.1 & 4.76 & \underline{81.1} & 3.23 & 76.9 & 4.99 \\
Seedream 4.5~\citep{Seedream4-5} & \underline{77.4} & 10.0 & \underline{78.5} & 17.5 & \textbf{84.5} & 5.88 & \underline{75.8} & 20.0 & \underline{74.9} & 7.14 & \underline{81.2} & 9.68 & \underline{78.7} & 11.7 \\
\rowcolor{gray!10} Seedream 5.0~\citep{Seedream5-0} & 69.0 & \underline{15.0} & 72.7 & 32.5 & 76.8 & 8.82 & 74.3 & 28.6 & 73.3 & 16.7 & 78.4 & 16.1 & 74.1 & 19.6 \\
\hline
\multicolumn{15}{l}{\textit{Open Source Models}} \\
\cdashline{1-15}
Bagel~\citep{deng2025bagel} & 59.9 & 0.00 & 63.8 & 0.00 & 66.5 & 0.00 & 72.5 & 2.86 & 66.5 & 0.00 & 68.5 & 0.00 & 66.3 & 0.476 \\
\rowcolor{gray!10} Bagel (think)~\citep{deng2025bagel} & 49.9 & 0.00 & 58.7 & 5.00 & 68.0 & 0.00 & 62.1 & 2.86 & 56.8 & 2.38 & 66.7 & 3.23 & 60.4 & 2.24 \\
DreamOmni~\citep{xia2025dreamomni} & 44.3 & 0.00 & 62.9 & 7.50 & 69.1 & 0.00 & 61.8 & 11.4 & 66.8 & 0.00 & 73.0 & 0.00 & 63.0 & 3.15 \\
\rowcolor{gray!10} FLUX.2 dev~\citep{flux-2-2025} & 60.0 & 0.00 & 67.9 & 7.50 & 71.3 & 5.88 & 70.2 & 11.4 & 61.4 & 2.38 & 76.2 & 3.23 & 67.8 & 5.07 \\
ICEdit~\citep{zhang2025icedit} & 49.7 & 5.00 & 59.3 & 0.00 & 63.9 & 0.00 & 60.9 & 2.86 & 57.3 & 0.00 & 67.7 & 0.00 & 59.8 & 1.31 \\
\rowcolor{gray!10} ICEdit (MoE)~\citep{zhang2025icedit} & 31.6 & 0.00 & 15.8 & 0.00 & 30.7 & 0.00 & 35.9 & 0.00 & 31.5 & 0.00 & 33.1 & 0.00 & 29.8 & 0.00 \\
OmniGen~\citep{xiao2025omnigen} & 60.7 & 0.00 & 63.7 & 0.00 & 59.2 & 0.00 & 58.7 & 0.00 & 63.5 & 0.00 & 63.1 & 0.00 & 61.5 & 0.00 \\
\rowcolor{gray!10} Qwen-Edit-2511~\citep{wu2025qwenimage} & \textbf{78.6} & 5.00 & \underline{78.1} & 17.5 & 80.2 & 5.88 & \underline{77.2} & 17.1 & \underline{77.4} & 0.00 & 77.3 & 16.1 & \underline{78.1} & 10.3 \\
Step-1x~\citep{han2025stepx} & 53.0 & 2.50 & 65.8 & 2.50 & 75.5 & 0.00 & 69.1 & 8.57 & 70.9 & 7.14 & 73.4 & 0.00 & 68.0 & 3.45 \\
\rowcolor{gray!10} Step-1x (think)~\citep{han2025stepx} & 56.6 & 0.00 & 64.9 & 2.50 & 75.1 & 2.94 & 68.4 & 5.71 & 71.6 & 0.00 & 74.6 & 0.00 & 68.5 & 1.86 \\
Step-1x (think+reflect)~\citep{han2025stepx} & 56.5 & 0.00 & 68.4 & 5.00 & 76.3 & 2.94 & 66.9 & 5.71 & 71.0 & 2.38 & 72.4 & 0.00 & 68.6 & 2.67 \\
\hline
\end{tabular}%
}
\label{tab:benchmark_qa_overall}
\end{table*}

Tables \ref{tab:benchmark_rc_overall} and \ref{tab:benchmark_qa_overall} present the full leaderboard. GPT-Image-2 is the strongest system overall, reaching 42.7\% strict accuracy with the highest reasoning score of 79.1 and near saturated consistency at 95.1. Nano Banana 2 follows at 39.2\% strict accuracy, scoring 76.2 in reasoning and 94.6 in consistency. Nano Banana Pro achieves 35.0\% accuracy while securing the best overall consistency score of 96.3. No evaluated system exceeds 50\% strict accuracy, confirming that benchmark defined geoscience visual reasoning remains far from solved even for frontier closed source systems.

The gap to open source models is large. The best open source system, Qwen-Edit-2511, reaches 10.3\% strict accuracy, which is roughly four times lower than the top closed source model. Five of the eleven open source systems score below 2\%, and OmniGen scores exactly 0\%. This spread is far wider than typical natural image editing benchmarks report, indicating that GeoR-Bench probes capabilities that open source editing models currently lack.

\subsection{Bottleneck Analysis}
A consistent pattern emerges from the component decomposition. Quality clusters tightly across the leaderboard. Almost every system lands in the 60 to 80 range, led by FLUX.2 Pro at 79.1 and followed closely by three models within two points. Consistency is more spread out, but the top seven systems all achieve scores above 90. Reasoning, by contrast, spans from 4.8 to 79.1 and covers more than an order of magnitude across the same model set. Reasoning is therefore the dimension that actually discriminates geoscience competence, as the other evaluation components saturate long before it does.

Two models make this concrete. FLUX.2 Pro holds the highest quality score in the table yet produces correct scientific transformations on only 5.6\% of samples because its reasoning is mediocre at 42.6 and its consistency is weak at 61.3. Qwen-Edit-2511 reaches a quality level of 78.1 to perform on par with frontier closed source systems and surpass Nano Banana 2, but its reasoning is capped at 48.9, which collapses its strict accuracy down to 10.3\%. These are precisely the visually plausible but scientifically wrong failures.

\begin{figure*}[t]
    \centering
    \includegraphics[width=\textwidth]{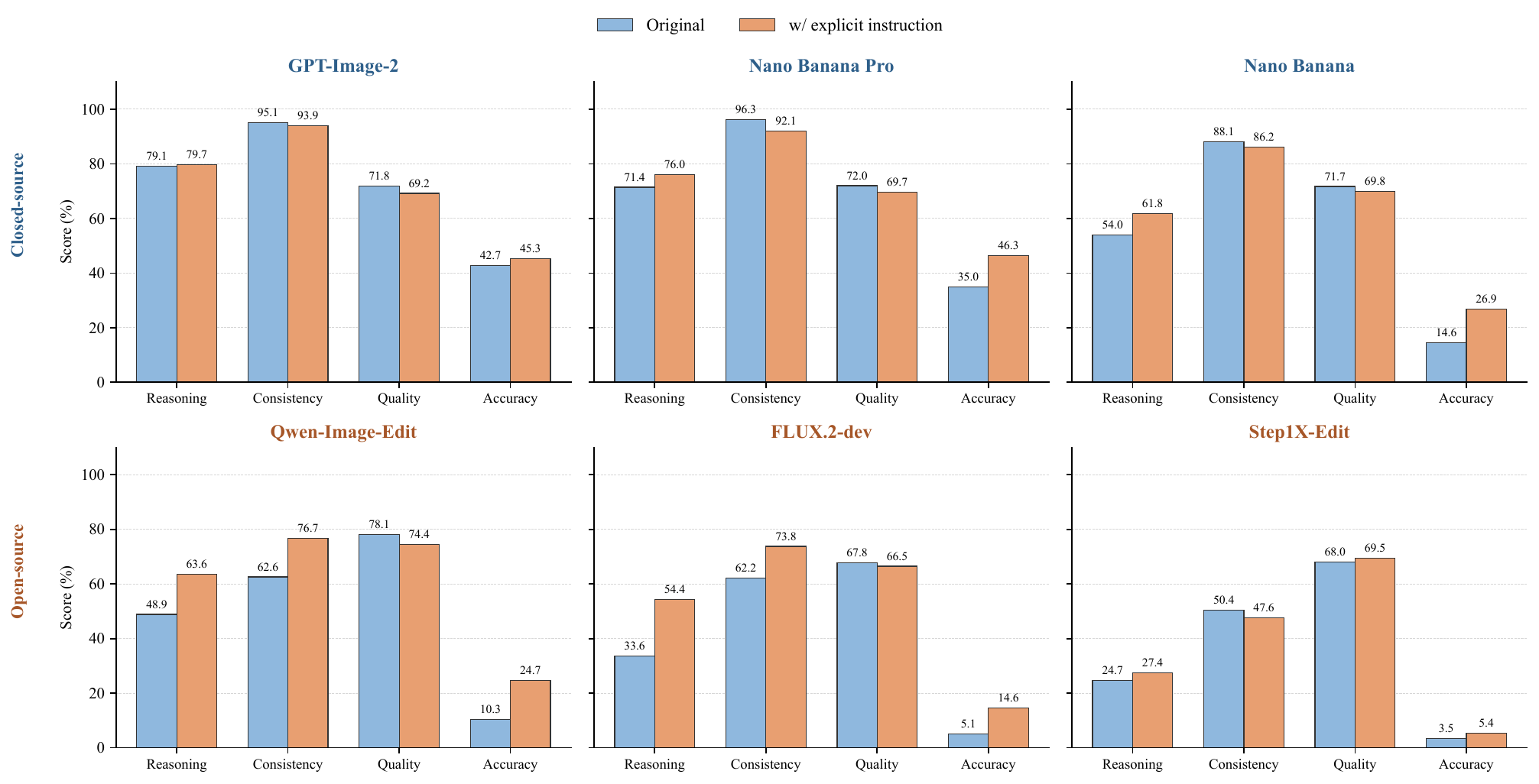}
    \caption{Comparison of model performance across different prompt settings.}
    \label{fig:compare}
\end{figure*}
\subsection{Category-level Analysis}
Category level accuracy in Table \ref{tab:benchmark_qa_overall} reveals strong domain asymmetries. Hydrology is the most tractable slice. Nano Banana Pro reaches 57.5\% strict accuracy here, and three closed source systems exceed 50\%. This is consistent with hydrologic task types being grounded in recurring temporal satellite series such as oxbow lakes and wetland fluctuation, where strong appearance priors align with correct geoscience behavior. The cryosphere category is similarly amenable, where GPT-Image-2 achieves 54.3\% strict accuracy since glacier retreat and snowline shifts exhibit clear directional signals.

Atmosphere and ocean dynamics form the hardest slice. Even GPT-Image-2 reaches only 35.3\% here, and most systems fall below 10\%. These tasks, including cyclone generation, transposition, and monsoon wet to dry transitions, require reasoning about rotational sense, the Coriolis effect, and circulation structure, none of which can be spoofed by surface visual priors. Geomorphology is also surprisingly demanding at the top, where the best performance reaches only 25.0\%, because delta progradation, dune migration, and oxbow cutoff are process based. The correct answer is determined by sediment transport and channel geometry rather than scene appearance. GIS and spatial geometry together with crustal science form a middle regime. GPT-Image-2 performs strongly by scoring 52.4 in spatial tasks and 41.9 in crustal science, and Nano Banana 2 follows closely with scores of 45.2 and 48.4 in these respective domains. However, weaker systems fall below 10\% in these categories due to the combined demands of viewpoint, scale, and symbol convention.

\subsection{Open-source Landscape}
Within open source systems we observe three additional patterns. First, Qwen-Edit-2511 is a clear leader with a reasoning score of 48.9 and a quality score of 78.1, making it the only open source system to break 10\% strict accuracy. Second, adding a chain of thought mechanism does not uniformly help. Step-1x with a think and reflect configuration reaches a reasoning score of 21.3 and an accuracy of 2.67\%, which falls marginally below the base Step-1x reasoning score of 24.7 and accuracy of 3.45\%. In contrast, Bagel benefits from explicit reasoning steps by improving its reasoning score from 12.7 to 18.7. The value of explicit reasoning steps is therefore dependent on the architecture rather than the specific task. Third, the Mixture of Experts variant of ICEdit is an outlier at the bottom of the leaderboard with a reasoning score of 7.1, consistency of 18.2, and quality of 29.8. This performs substantially worse than the base ICEdit scores of 13.2, 31.1, and 59.8 respectively. The quality collapse is especially informative because it shows the Mixture of Experts variant is not just reasoning weak but also generation unstable on geoscience inputs. Together these observations suggest that current open source editing systems remain architecturally tuned for natural image edits and require dedicated alignment for scientific content.

\begin{table}[t]
\centering
\small
\caption{Alignment between automated evaluation and expert ratings. Lower is better.}
\label{tab:alignment}
\begin{tabular}{lcccccc}
\toprule
& \multicolumn{2}{c}{Reasoning} & \multicolumn{2}{c}{Consistency} & \multicolumn{2}{c}{Image Quality} \\
Judge model 
& MAE$\downarrow$ & STD$\downarrow$ 
& MAE$\downarrow$ & STD$\downarrow$ 
& MAE$\downarrow$ & STD$\downarrow$ \\
\midrule

Qwen3-VL-235B~\cite{qwen3technicalreport}
& 0.180 & 0.350
& 0.123 & 0.352
& 0.094 & 0.281 \\

GPT-5~\cite{singh2025openai}
& 0.152 & 0.323
& 0.141 & 0.402
& 0.082 & 0.248 \\

Gemini-3-Flash~\cite{Gemini3-Flash}
& \textbf{0.119} & \textbf{0.283}
& \textbf{0.095} & \textbf{0.305}
& 0.072 & 0.215 \\

Q-Align~\cite{wu2023qalign}
& -- & --
& -- & --
& \textbf{0.063} & \textbf{0.158} \\

\bottomrule
\end{tabular}
\end{table}

\subsection{Abliation Study}
\paragraph{Instruct Prompt} Many GeoR-Bench prompts intentionally require implicit geoscience reasoning by asking the model to evolve a delta rather than explicitly instructing it to add distributary channels seaward. To test how much models benefit from explicit guidance, we rewrite a matched subset of prompts into procedural instructions while keeping the input and target output fixed. Figure~\ref{fig:compare} shows a clear asymmetry. Open source systems gain the most from explicit instructions. Qwen-Edit-2511 improves from 10.3\% to 24.7\% strict accuracy, which is a 14.4 point absolute gain. FLUX.2-dev increases from 5.1\% to 14.6\% for a 9.5 point gain, and Step1X-Edit sees a modest 1.9 point improvement. This aligns with the hypothesis that open source systems rely on overt decomposition rather than implicit geoscience priors. Closed source systems gain unevenly. GPT-Image-2 improves only modestly from 42.7\% to 45.3\%, gaining 2.6 points, which is consistent with its implicit reasoning already being strong. In contrast, Nano Banana Pro gains 11.3 absolute points to reach 46.3\%. This is the highest accuracy observed in any configuration in this paper and sits slightly above the explicit setting score of 45.3\% achieved by GPT-Image-2. The implication is that Nano Banana Pro has latent geoscience capability that is activated by explicit decomposition, while GPT-Image-2 is already exercising its reasoning more fully under implicit prompts.  

\paragraph{Judge Model}
We assess the agreement between automated judging and expert ratings on a 48 sample subset spanning the benchmark categories. Table~\ref{tab:alignment} reports the mean absolute error and standard deviation between the automated scores and averaged expert judgments. Gemini-Flash achieves the best comprehensive alignment across the evaluation suite. It records the lowest error in reasoning with a mean absolute error of 0.119 and the best consistency alignment at 0.095, effectively outperforming both GPT-5 and Qwen3-VL-235B. 
The dedicated Q-Align model achieves the lowest error for the quality dimension at 0.063 so during the image quality assessment, it performs best. These results validate our configuration and strongly support the use of structured scoring points and task specific evaluation prompts.

\section{Conclusion}
\label{sec:conclusion}
We present GeoR-Bench, a benchmark that evaluates geoscience intelligence through structured image reasoning and generation tasks. This dataset stems from the core observation that geoscience images inherently carry strict scientific meaning, which dictates that proper evaluation must extend far beyond standard visual realism. By grounding the evaluation in curated geoscience imagery across 440 distinct samples spanning six categories and 24 task types, GeoR-Bench measures the geoscience intelligence of the multi-modal models instead of the aesthetics. Our framework jointly assesses reasoning, consistency, and quality to reveal significant limitations in the current multimodal landscape, the geoscience visual reasoning. Which causes no evaluated models achieves a 50\% strict accuracy threshold. Ultimately, we hope GeoR-Bench guides future research toward physically grounded and scientifically faithful image generation for critical Earth system applications.

\bibliographystyle{unsrtnat}
\bibliography{references}

\end{document}